\def\paperID{5903}
\def\confName{CVPR}
\def\confYear{2024}

\def\paperTitle{Towards Learning a Generalist Model for Embodied Navigation}

\def\authorBlock{
    Duo Zheng$^{1,2}$\thanks{Equal contribution.} \qquad
    Shijia Huang$^1$\footnotemark[1] \qquad
    Lin Zhao$^{3}$\thanks{Lin Zhao was a research assistant at the Centre for Perceptual and Interactive Intelligence (CPII) under the InnoHK.} \qquad
    Yiwu Zhong$^1$ \qquad
    Liwei Wang$^1$\thanks{Corresponding author.} \qquad \\
    $^1$The Chinese University of Hong Kong \quad
    $^2$Shanghai AI Laboratory\\
    $^3$Centre for Perceptual and Interactive Intelligence \\
    {\tt\small \{dzheng23, sjhuang, lwwang\}@cse.cuhk.edu.hk}
}

\newif\ifreview 
\newif\ifarxiv \newcommand{\arxiv}{\arxivtrue}
\newif\ifcamera 
\newif\ifrebuttal

\arxiv
\pdfoutput=1
\documentclass[10pt,twocolumn,letterpaper]{article}
\ifreview \usepackage[review]{cvpr} \fi
\ifarxiv \usepackage[pagenumbers]{cvpr} \fi
\ifrebuttal \usepackage[rebuttal]{cvpr} \fi
\ifcamera \usepackage{cvpr} \fi

\usepackage{graphicx}	
\usepackage{amsmath}	
\usepackage{amssymb}	
\usepackage{booktabs}
\usepackage{times}
\usepackage{microtype}
\usepackage{epsfig}
\usepackage[table,xcdraw,dvipsnames]{xcolor}
\usepackage{caption}
\usepackage{float}
\usepackage{placeins}
\usepackage{color, colortbl}
\usepackage{stfloats}
\usepackage{enumitem}
\usepackage{tabularx}
\usepackage{xstring}
\usepackage{multirow}
\usepackage{xspace}
\usepackage{url}
\usepackage{subcaption}
\usepackage{xcolor}
\usepackage[hang,flushmargin]{footmisc}

\usepackage{rotating}
\usepackage{marvosym}
\usepackage{pifont}

\ifcamera \usepackage[accsupp]{axessibility} \fi

% ...

\newcommand{\modelname}{\textit{NaviLLM}}
\newcommand{\para}[1]{\vspace{0.5ex}\noindent\textbf{#1}}

\ifarxiv  \fi

\newcommand{\R}[1]{{%
    \textbf{%
        \ifstrequal{#1}{1}{\textcolor{red}{R#1}}{%
        \ifstrequal{#1}{2}{\textcolor{blue}{R#1}}{%
        \ifstrequal{#1}{3}{\textcolor{magenta}{R#1}}{%
        \ifstrequal{#1}{4}{\textcolor{teal}{R#1}}{%
                           \textcolor{cyan}{R#1}%
        }}}}%
    }%
}}
\usepackage{xr-hyper}
\usepackage{makecell}
\makeatletter
\newcommand*{\addFileDependency}[1]{
  \typeout{(#1)}
  \@addtofilelist{#1}
  \IfFileExists{#1}{}{\typeout{No file #1.}}
}

\makeatother

\definecolor{cvprblue}{rgb}{0.21,0.49,0.74}
\usepackage[pagebackref,breaklinks,colorlinks,citecolor=cvprblue]{hyperref}
\usepackage[capitalize]{cleveref}
\crefname{section}{Sec.}{Secs.}
\crefname{table}{Table}{Tables}
\crefname{figure}{Fig.}{Figs.}

\begin{document}
%% TITLE
\title{\paperTitle}
\author{\authorBlock}
\maketitle

\begin{abstract}
% Abstract goes here.

Building a generalist agent that can interact with the world is the intriguing target of AI systems, thus spurring the research for embodied navigation, where an agent is required to navigate according to instructions or respond to queries. 
Despite the major progress attained, previous works primarily focus on task-specific agents and lack generalizability to unseen scenarios. 
Recently, LLMs have presented remarkable capabilities across various fields, and provided a promising opportunity for embodied navigation.
Drawing on this, we propose the first generalist model for embodied navigation, \modelname.
It adapts LLMs to embodied navigation by introducing schema-based instruction.
The schema-based instruction flexibly casts various tasks into generation problems, thereby unifying a wide range of tasks.
This approach allows us to integrate diverse data sources from various datasets into the training, equipping \modelname~with a wide range of capabilities required by embodied navigation.
We conduct extensive experiments to evaluate the performance and generalizability of our model.
The experimental results demonstrate that our unified model achieves state-of-the-art performance on CVDN, SOON, and ScanQA. 
Specifically, it surpasses the previous stats-of-the-art method by a significant margin of 29\% in goal progress on CVDN. 
Moreover, our model also demonstrates strong generalizability and presents impressive results on unseen tasks, \eg embodied question answering and 3D captioning.
Our code is available at \url{https://github.com/LaVi-Lab/NaviLLM}.

\end{abstract}

\section{Introduction}
\label{sec:intro}

% Large Language Models (LLM) \cite{NEURIPS2020_1457c0d6, chowdhery2022palm, touvron2023llama} have revolutionized the ability to understand and generate text, and recent advancements \cite{liu2023visual, yang2023dawn} have even expanded their capabilities to handle visual inputs.
% However, the potential to fully harness AI's potential in interacting with and learning from an interactive environment remains an unresolved challenge. 
\begin{figure}[] %H为当前位置，!htb为忽略美学标准，htbp为浮动图形
\centering %图片居中
\includegraphics[width=0.46\textwidth]{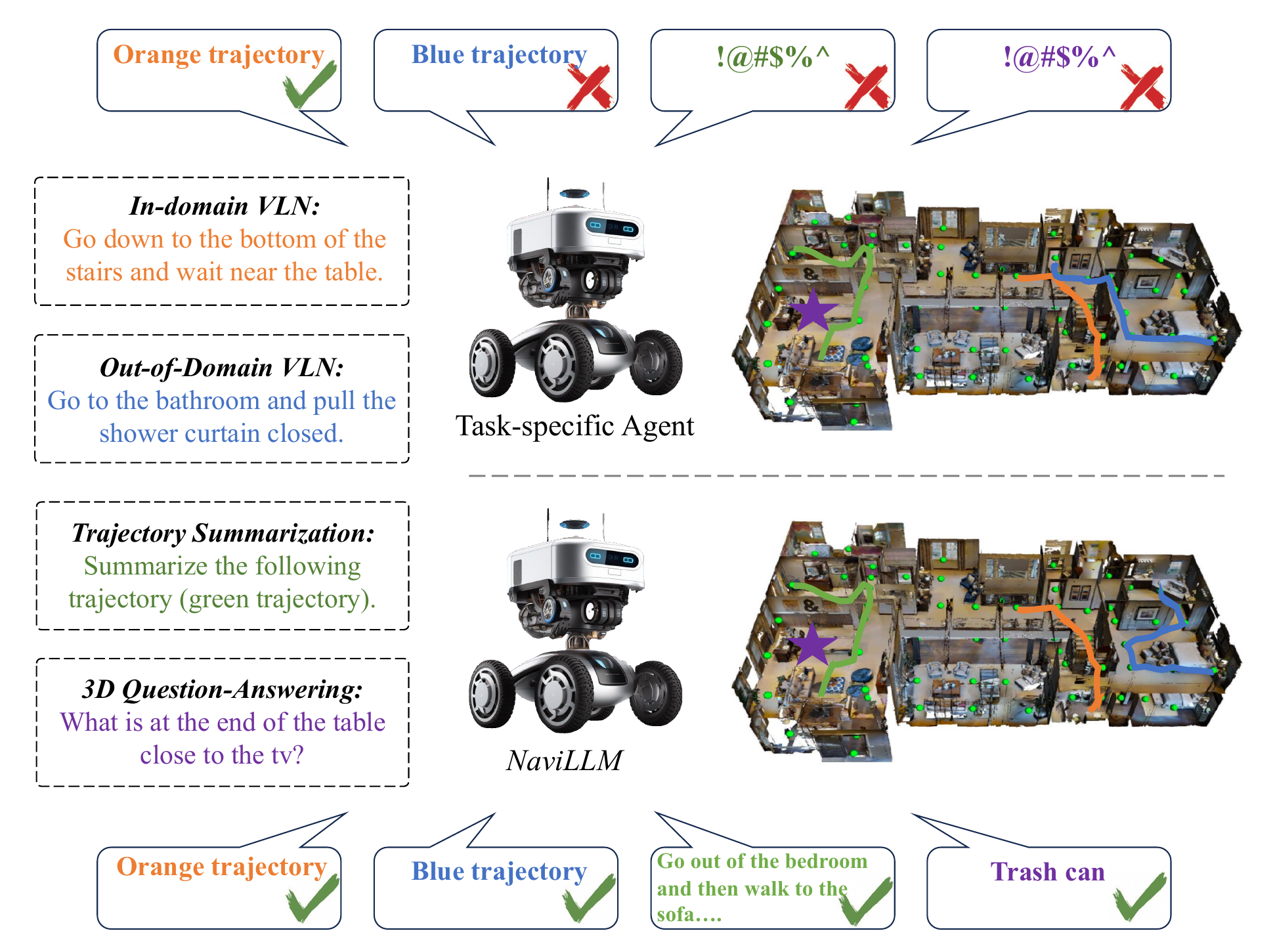} %插入图片，[]中设置图片大小，{}中是图片文件名
\vspace{-0.1cm}
\caption{Comparison between previous methods and ours. Previous methods learn task-specific navigation agents, suffer from a low success rate for out-of-domain VLN, and fall short when facing unseen tasks (\eg, QA and summarization). The different colors are used to represent different examples. 
For instance, orange represents an example from In-domain VLN. Our \modelname~not only excels in diverse tasks required by embodied navigation, but also demonstrates promising generalizability even on unseen tasks.}
\label{fig:intro}
\vspace{-0.2cm}
\end{figure}
The pursuit of artificial intelligence has long been driven by the desire to construct agents that are capable of acquiring knowledge through interacting with the physical world, akin to the complex interaction processes exhibited by humans. 
This has led to the emergence of embodied navigation \cite{anderson2018evaluation, savva2019habitat, ai2thor}, where an agent located in 3D environment is required to navigate according to various forms of instructions and provide textual responses based on user queries.

A wide spectrum of tasks have been introduced for embodied navigation, ranging from vision-language navigation that follows step-by-step instructions \cite{anderson2018vision, ku2020room} or coarse-grained directives \cite{das2018embodied, qi2020reverie, zhu2021soon}, to the tasks guided by interactions between humans and agents \cite{de2018talk, thomason2020vision}, and even to embodied question answering through proactive exploration \cite{embodiedqa,wijmans2019embodied}.
To tackle these tasks, a myriad of methodologies have been explored in the past, with some notable approaches leveraging pre-training techniques \cite{majumdar2020improving, li2023vlnsig, hao2020towards, guhur2021airbert}, data augmentation \cite{chen2022learning, fried2018speaker, huang2019transferable}, and memory structures \cite{Hong_2021_CVPR, NEURIPS2021_2e5c2cb8}, etc.  
These models, while demonstrating considerable proficiency in their specific tasks, unfortunately lack generalization across diverse scenarios. %lack the flexibility to adapt and perform equivalently across diverse scenarios.
This naturally raises a question: {\it Can we train a generalist model that is generalizable to many embodied navigation tasks?}

% Yiwu: suggestions for figures

The advancement of Large Language Models (LLMs) has provided a promising opportunity to construct a generalist model for embodied navigation.
In recent years, LLMs \cite{NEURIPS2020_1457c0d6, chowdhery2022palm, touvron2023llama} have demonstrated human-like capability for text understanding and text generation.
Given such impressive performance, numerous works \cite{dai2023instructblip, liu2023visual, zhu2023minigpt4, Qwen-VL} have pioneered adapting LLMs for vision-language tasks through fine-tuning on a variety of data sources.
Beyond the image domain, LLMs have exhibited remarkable generalizability in other domains, such as video understanding~\cite{li2023videochat}, 3D understanding~\cite{hong20233dllm}, and robotic manipulation~\cite{brohan2022rt, shridhar2023perceiver, brohan2023rt2}.
% While being predominantly used in natural language understanding and generation, 
However, the potential of adapting LLMs to embodied navigation tasks remains largely unexplored.

In this work, we aim to learn a generalist model for embodied navigation by adapting LLMs. The main challenge lies in how to unify a wide range of tasks in a single model. To address the challenge, our key idea is to cast all task learning into generative modeling, with the help of pretrained LLMs.
Specifically, we propose schema-based instruction and design a series of schemas (\eg, descriptions of tasks, visual observation, and navigation history), based on the characteristics of embodied tasks. These schemas are flexible to cast various vision-centric tasks into generation problems.
For example, we can effortlessly convert vision-language navigation into the generation of movement direction, and convert object localization into the generation of object IDs.
% We construct the prompt by combining related schemas. Each schema adheres to a unified format, yet varies across different data sources, thus enabling flexible adaption for a variety of tasks.
% More specifically, the \textit{Observation} schema is an alternating sequence of visual and textual representations, where the ID tokens are prepended to the front of corresponding visual representations.
% This enables us to transform vision-centric tasks such as Visual Language Navigation and Object Localization into tasks that predict the IDs of movement directions and objects, respectively.
% By leveraging these carefully designed schemas, we are able to transform vision-centric tasks into generation problems, such as converting Visual Language Navigation and Object Localization into predicting movement direction and object IDs, respectively.
Benefitting from this design, we are able to train a unified model on the data collected for diverse tasks, thereby enabling our model to address a wide spectrum of tasks, ranging from vision-language navigation and object localization, to 3D question answering, trajectory summarization, embodied question answering.
Therefore, our approach significantly mitigates the problem of data scarcity and empowers the model to understand instructions of varying formats and granularities, thereby enabling a suite of capabilities to interact with the 3D environment.

% Yiwu: suggestion for empirical results
We train \modelname~on a combined dataset covering diverse embodied tasks (CVDN, SOON, R2R, REVERIE, ScanQA, LLaVA-23k, and augmented data for R2R and REVERIE), and conduct extensive experiments to evaluate the competencies and generalizability of \modelname. 
With only a single model, \modelname~has achieved \textbf{new state-of-the-art} results simultaneously on multiple benchmarks, \ie CVDN \cite{thomason2020vision}, SOON \cite{zhu2021soon}, and ScanQA \cite{azuma2022scanqa}, and demonstrated comparable performance to latest models on R2R \cite{anderson2018vision} and REVERIE \cite{qi2020reverie}. 
Notably, our model achieves a relative improvement of \textbf{29\%} over the previous state-of-the-art on the CVDN benchmark.
Further, we evaluate the generalizability of our method by excluding CVDN, SOON, and REVERIE from the training data, respectively.
Our method outperforms baseline methods on all tasks, significantly improving the Success Rate by \textbf{136\%} on SOON. % compared DUET \cite{Chen_2022_CVPR} trained on REVERIE.
Moreover, we observe that our model presents an impressive capability for unseen tasks, \eg embodied question answering and 3D captioning.
Collectively, the results from these experiments not only attest to the generalization capability of the model but also highlight the significant potential of our approach in learning a generalist model for embodied navigation.
% Furthermore, we observe that our method shows significantly greater improvement on the tasks that have complex instructions, compared to those with simple instructions. This demonstrates that the knowledge inherited from LLM greatly improves the comprehension of complex instructions.
% We implement zero-shot transfer experiments on the MP3D-EQA dataset, during which our model demonstrated 33.5\%  SPL and 44\% Accuracy. 
% Additionally, we conduct held-out experiments to empirically validate the model's proficiency in handling tasks not encountered during the training phase. 

% Yiwu: suggestions for contributions
Our contribution could be summarized as follows:
\begin{itemize}[leftmargin=*]
\setlength{\itemsep}{0pt}
\setlength{\parsep}{0pt}
\setlength{\parskip}{0pt}
% \item We propose \modelname, a generalist model for embodied navigation. It is built upon the foundation of LLM and leverages the extensive world knowledge inherited from LLM, enabling it to generalize across various tasks.
% \item We incorporate eight datasets into the training of our model to mitigate the problem of data scarcity, and adapt our model to a series of tasks.
% \item Our unified model achieves SoTA performance on CVDN, SOON, and ScanQA, while also delivering competitive results on R2R and REVERIE. This underscores the effectiveness of our approach across diverse tasks.
% \item Our model delivers 33.5\% SPL and 44\% Accuracy in zero-shot transfer experiments on MP3D-EQA and shows proficiency in held-out experiments on CVDN, SOON, and REVERIE, demonstrating robust generalizability.
\item We propose the first generalist model for embodied navigation, namely \modelname, enabling a wide spectrum of capabilities required for embodied navigation.
\item We unify various tasks in a single model by adapting LLM and introducing schema-based instruction. By doing this, our model can harness data sources from diverse datasets.
\item Our single model achieves SoTA results on CVDN, SOON, and ScanQA, with a significant margin of \textbf{29\%} compared to the previous SoTA on CVDN. Furthermore, it also exhibits strong generalizability on unseen tasks.
\end{itemize}

% Despite these remarkable efforts, there remains significant room for improvement of VLN models.
% Distinct tasks pose unique challenges. For instance, Room-to-Room (R2R) necessitates an agent's ability to track its current state, while REVERIE requires the agent to have capabilities for environment exploration and object grounding. Cooperative Vision-and-Dialogue Navigation (CVDN), on the other hand, demands the model's proficiency in dialogue comprehension.

% This task inherently requires the learning of situated representations, which involve understanding contextual information and making inferences accordingly.

% To insert a figure: \input{figs/template}
% Or table: \input{tables/template}

\section{Related Work}
\label{sec:related}

\subsection{Vision-Language Navigation}
% notes from Yiwu
% 1. terms of tasks vs. datasets vs. capabilities vs. approaches
% 2. missing information
Vision-language navigation has been extensively explored in the last few years, including a variety of tasks that require different aspects of embodied capabilities. % There is a wide variety of vision-and-language navigation tasks that necessitate a suite of capabilities.
R2R \cite{anderson2018vision} requires the agent to navigate the rooms step by step, following fine-grained instructions, while Thomason \textit{et al.} \cite{thomason2020vision} introduce Cooperative Vision-and-Dialog Navigation (CVDN) which demands the agent to navigate based on a dialog history.
%, thereby demanding the modeling for history.
Beyond navigation, SOON \cite{zhu2021soon} and REVERIE \cite{qi2020reverie} additionally require the agent to localize the objects queried in instructions.
EQA \cite{embodiedqa, wijmans2019embodied} emphasizes the ability to answer questions in a 3D environment by actively exploring the environment.
To tackle these tasks, significant efforts have been invested.
However, previous approaches \cite{hao2020towards, NEURIPS2021_2e5c2cb8, Hong_2021_CVPR, Chen_2022_CVPR, Hwang_2023_CVPR, Gao_2023_CVPR, Li_2023_CVPR} primarily focus on designing specialist models for individual tasks, and these models often struggle to generalize and transfer to other tasks. %, lacking in generalization capabilities.
Our method, as an embodied generalist, addresses these tasks simultaneously via a single model and demonstrates strong generalization capabilities.

The work closely related to ours is MT-RCM \cite{wang2020environment}, a multi-task model, designed to alleviate overfitting to specific datasets.
On the other hand, our method primarily leverages the LLMs to enhance generalizability, and LLMs have been adapted to a broader range of tasks and datasets.
Recently, some work \cite{dorbala2022clip, zhou2023navgpt} has also explored the generalization and transferability of agents by utilizing off-the-shelf foundation models. In contrast to their approach, our method focuses on building a unified, end-to-end embodied learner, rather than a pipeline chaining up multiple independent models.

\subsection{Multimodal Instruction Tuning}
Large Language Models (LLMs) \cite{NEURIPS2020_1457c0d6, chowdhery2022palm, touvron2023llama} have revolutionized text understanding and text generation. Recent advancements \cite{liu2023visual, yang2023dawn} further expanded their capabilities to digest visual inputs.
For example, multimodal instruction tuning methods \cite{dai2023instructblip, liu2023visual, zhu2023minigpt4, Qwen-VL, li2023videochat} have been widely proposed for 2D images \cite{dai2023instructblip} or videos \cite{li2023videochat}. 
One of the distinctions between our work and other multi-modal LLMs is that, our work is designed for embodied AI, including navigation and 3D understanding, which previous works don't consider.
More recently, Hong et al. \cite{hong20233dllm} introduced 3D-LLM by adapting LLMs to 3D data.
However, it does not address the problem of calibrating LLMs for embodied navigation that requires the ability of sequential decision-making. 
% While \cite{hong20233dllm} has made strides with the 3D-LLM to handle 3D representations, it does not fully address navigation aspects like decision-making and history information modeling. 
% Our method distinguishes itself from existing approaches by not only learning navigation capabilities but also demonstrating proficiency in 3D-related tasks. 

\subsection{Large Language Models as Embodied Agents}
Two lines of work have emerged in the exploration of integrating LLMs into embodied tasks. 
The first line focuses on translating visual information into textual format, which is then processed by the LLMs to generate plans \cite{brohan2023can, huang2023inner, mu2023embodiedgpt, zhou2023navgpt}, landmarks \cite{shah2023lm}, or code \cite{liang2023code}. 
These methods leverage the inherent knowledge within the pre-trained, frozen LLMs. %, but do not directly fine-tune LLMs.
The second line \cite{brohan2022rt, shridhar2023perceiver, brohan2023rt2}, in contrast, directly fine-tunes LLMs on datasets that comprise action sequences of robotic manipulations. 
% facilitating a more harmonious integration of LLMs with physical environments.
Similar to the second line of methods, our work also fine-tunes LLMs, yet focuses on addressing various tasks in embodied navigation, rather than robotic manipulations.

% Our work falls into the second category but sets itself apart by focusing on VLN rather than low-level control manipulation tasks.
% However, these methods primarily concentrate on manipulation tasks involving low-level controls, while our work focuses on VLN, addressing a distinct aspect of embodied tasks.

\begin{figure*}[] %H为当前位置，!htb为忽略美学标准，htbp为浮动图形
\centering %图片居中
\includegraphics[width=1.\textwidth]{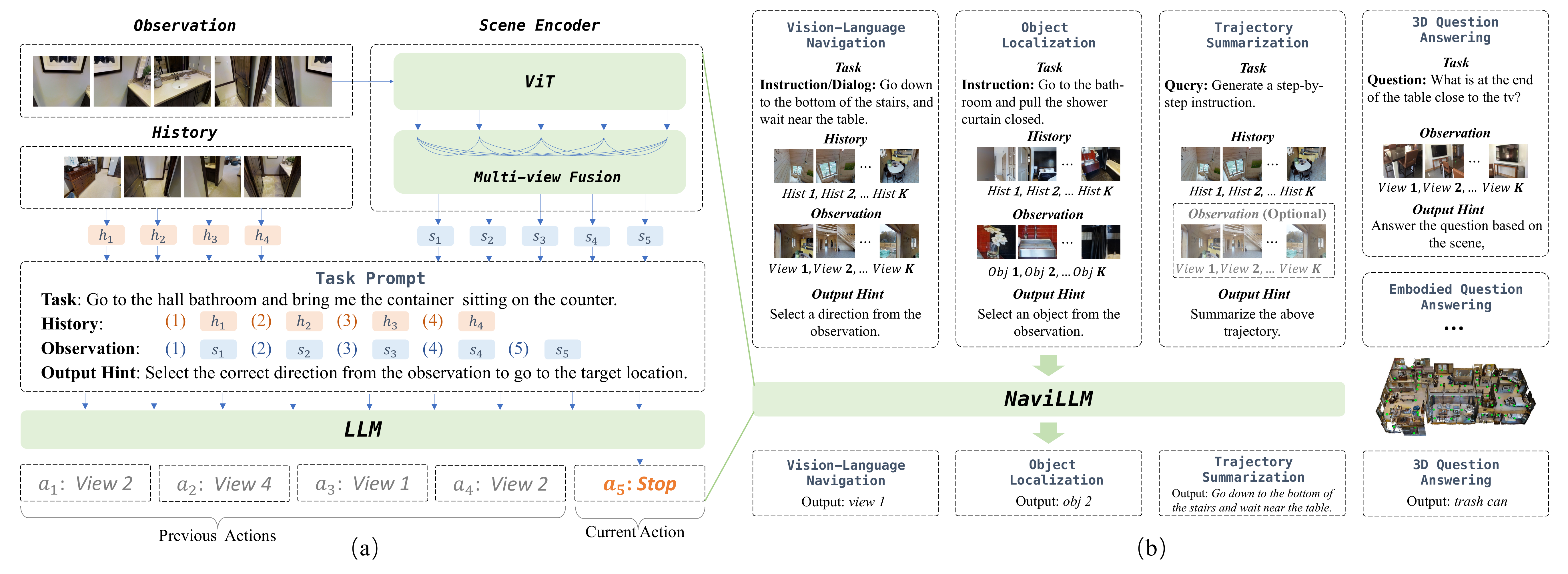} %插入图片，[]中设置图片大小，{}中是图片文件名
\caption{The overview of \modelname. The left figure presents the architecture and workflow of our model, while the right figure illustrates the schema-based instruction and multi-task learning process in our method.}
\label{fig:method}
\vspace{-0.2cm}
\end{figure*}

\section{Method}
\label{sec:method}
% In this section, we first provide the problem formulation of embodied navigation (\S \ref{sec:formulation}), and then introduce \modelname (\S \ref{sec:model}).
% Finally, we demonstrate how to apply multi-task learning to our method (\S \ref{sec:multi-task}).

% \subsection{Model Architecture}
% \label{sec:model_arch}
\subsection{Problem Formulation} 
\label{sec:formulation}
In embodied navigation, an embodied agent situated in the 3D environment is required to complete tasks described in natural language. 
% The agent aims to predict actions based on past trajectories and current observations. 
The agent leverages past trajectories and current observations to predict actions that enable task completion.
% The actions can either be a navigation move, the bounding box of an object, or a text response.
The actions encompass navigation moves, bounding boxes for objects, and textual responses.
% At each step $t$, the agent receives an observation $\mathcal{V}_t=\{(I_i, \theta_i, g_i)\}_{i=1}^n$ from multiple viewpoints. Here, each tuple $(I_i, \theta_i,g_i)$ corresponds to the image from a specific viewpoint, its associated angle, and the corresponding GPS information, respectively. 
% As the agent navigates through the environment, it maintains a history $\mathcal{H}_t$ to record the past visual observations. 
% The history is represented as $\mathcal{H}_t=\{(h_i)\}_{i=1}^t$, where each $h_i$ denotes a past observation and is updated at each step. 

% \subsection{Model Components} 
\subsection{NaviLLM} 
\label{sec:model}
\modelname\ is an embodied model grounded in LLM, comprising two modules, \ie, a scene encoder and an LLM.
As depicted in Figure \ref{fig:method} (a), the scene encoder takes the current visual observation as input, and transforms it into a series of scene representations (\S \ref{sec:scene_enc}).
Utilizing these scene representations, we construct various schemas for different tasks and these schemas serve as input for LLM, to produce the next action (\S \ref{sec:schema}).

\subsubsection{Scene Encoding} 
\label{sec:scene_enc}
% The scene encoder is responsible for extracting scene presentations given an observation from multiple viewpoints $\mathcal{V}=\{(I_i, \theta_i, g_i)\}_{i=1}^n$.
% Each triplet $(I_i, \theta_i,g_i)$ corresponds to an image from a specific viewpoint, its associated angle, and the corresponding GPS information, respectively.
The scene encoder extracts scene presentations given an observation composed of a set of images $\{I_i\}_{i=1}^n$, with each image representing a unique viewpoint.
The visual encoder initially extracts visual features for each individual image via a Vision Transformer (ViT) \cite{dosovitskiy2020image}. These features from different viewpoints are then integrated via a multi-view fusion process, yielding scene representations $\{s_i\}_{i=1}^n$.

\noindent \textbf{Visual Feature Extraction.} A pre-trained ViT is utilized to extract visual features from images.
Specifically, given an image $I_i$, it is first divided into a sequence of patches, with a special [CLS] token appended at the beginning. The sequence is then fed into the transformer network of the ViT.
Finally, the last hidden states of the [CLS] token serve as visual features, which could be denoted as:

\begin{equation}
f_i=\text{ViT}(I_i) \quad \text{for } i = 1, 2, ..., n,
\end{equation}
where $f_i$ is the visual features of the $i$-th viewpoint.

% \para{Multi-View Fusion.}
% Upon acquiring the visual features, multi-view fusion is performed to model complex interdependencies among different viewpoints.
% Firstly, the visual features $f_i$ are combined with the encoding of its corresponding angle information $\theta_i$, which could be denoted as:
% \begin{equation}
% c_i = f_i + \phi^{angle}(\theta_i),
% \end{equation}
% where $c_i$ is the combined vectors and $\phi^{angle}$ is a linear layer.
% Then the combined vectors $\{c_i\}_{i=1}^n$ from all viewpoints are fed into a transformer encoder, to learn the spatial relationships between different viewpoints, formulated as:
% \begin{equation}
% \{h_i\}_{i=1}^n=\text{Transformer-Encoder}(\{c_i\}^n_{i=1}),
% \end{equation}
% \noindent where $\{h_i\}_{i=1}^n$ are the last hidden states of the transformer encoder.
% Lastly, the last hidden states are added with the encodings of their corresponding GPS information, to obtain the scene representations:
% \begin{equation}
% s_i = h_i + \phi^{GPS}(g_i),
% \end{equation}
% where $h_i$ and $s_i$ are the last hidden states and scene representation for the $i$-th viewpoint, and $\phi^{GPS}$ is a linear layer.
\para{Multi-View Fusion.}
Upon acquiring the visual features, multi-view fusion is performed to model complex interdependencies among different viewpoints.
The image features $\{f_i\}_{i=1}^n$ for all viewpoints are fed into a transformer encoder, to learn the spatial relationships between different viewpoints, formulated as:
\begin{equation}
\{s_i\}_{i=1}^n=\text{Transformer-Encoder}(\{f_i\}^n_{i=1}),
\end{equation}
where $s_i$ is the scene representation for the $i$-th viewpoint.
To enhance the scene representations, we also incorporate the angle and GPS information of each view into the scene encoding. We omit these details for brevity.

\subsubsection{Schema-Based Instruction} 
\label{sec:schema}
% The LLM processes alternating visual and textual representations as input, and then utilizes its extensive comprehension capabilities and world knowledge to generate the appropriate action pertaining to a given task.

% Version of Duo Zheng
% The traditional schema-based representation in multi-turn dialog \cite{rastogi2020towards, chen2020schema, lee2021dialogue} serves as an effective means of generalizing to new tasks.
% In this work, we expand this concept to a multimodal paradigm, and carefully design schema-based instruction that can accommodate multimodal inputs.
% It flexibly combines distinct schemas as input for LLM. 
% Each schema adheres to a unified format, yet varies across different data sources, thus enabling flexible adaption for a variety of tasks.
% % For embodied navigation, we could decompose the task input as the combination of the following schemas.

% Edited version by Yiwu
Schema-based instruction was proposed in the language models for multi-turn dialog \cite{rastogi2020towards, chen2020schema, lee2021dialogue}, serving as an effective way of generalizing to novel tasks.
In the context of LLMs, we extend schema-based instruction to multimodal modeling so that it can digest multimodal information. 
Our schema is designed to be a unified format that can adapt to different data sources and enable flexibility for a wide range of tasks.

\para{\textit{Task}.} It consists of a word sequence that the agent is expected to execute, which could manifest in various forms, such as a navigation instruction, an invisible object required to find, or a question raised by a user.
% To assist the LLM in comprehending various instructions, we also provide a brief explanation of the characteristics of the dataset here.

\para{\textit{Observation}.} This refers to the visual observation at the current location of the agent.
The observation schema consists of the scene representations $\{s_i\}_{i=1}^n$.
To distinguish representations between different views, we prepend each representation with an ID, denoted by
\begin{equation}
[Emb(1), {s}_1, ..., Emb(n), {s}_n ],
\end{equation}
where $Emb(i)$ is the embedding of the ID for the i-th view. 

\para{\textit{History}.} It records the sequence of past visual observations 
 upon the $t$-th step. 
This schema provides a temporal context that helps the agent understand its past trajectory within the environment and the visual feedback associated with each decision. 
Given history representations $\{h^i\}_{i=1}^t$, we prepend each representation with an ID to indicate the order of the past observations.
The History schema is constructed as
\begin{equation}
[Emb(1), h^1, ..., Emb(t), h^t],
\end{equation}
$Emb(i)$ is the embedding of the ID for the i-th step.

\para{\textit{Output Hint}.} The schema hints at the output information that the agent is expected to produce, \eg, an identifier for a desired viewpoint to move towards, an answer response to a question, or a summarization for a previous trajectory.
This schema helps the model understand how to generate actions that align with the task requirements.

% \para{Query} It provides a query that the model requires to respond to, concerning either the current scene or the agent's historical trajectory, e.g., answering a question or producing a summarization for a given trajectory.

% \para{Response} The Response schema, paired with the Query schema, prompts the model to provide meaningful feedback in the form of natural language based on its understanding of the current scene or the agent's history trajectory.

\subsection{Multi-task Learning}
\label{sec:multi-task}
% To endow our model with a suite of capabilities, we incorporate various tasks using multi-task learning, facilitated by the unified model architecture and schema design.

% \subsubsection{Task}
As illustrated in Figure \ref{fig:method} (b), we summarize the key tasks for embodied navigation and transform these tasks into generation problems using schema-based instruction. 
Then we can optimize our model on different tasks with a unified cross-entropy objective. We detail each task as follows.
% During training, we randomly sample data points from the unified dataset and optimize the model with a unified cross-entropy objective.
% Figure \ref{fig:method} (b) illustrates the input schemas and output for each task, and we detail each task below.

\para{Vision-Language Navigation (VLN)} requires the agent to navigate in 3D environment to accomplish a given task.
We present the schemas for VLN as follows:
% At time step $t$, it receives a task description, history representations $\{h^i\}_{i=1}^t$, scene representations of all reachable viewpoints under the current position $\{s^t_i\}_{i=1}^n$.
% At the step $t$, the model receives a \textit{task schema} (an instruction), an observation schema (scene representations of all reachable viewpoints under the current position $\{s^t_i\}_{i=1}^n$), a history schema (history representations $\{h^i\}_{i=1}^t$), and an output hint schema.
\begin{itemize}[leftmargin=*]
\setlength{\itemsep}{0pt}
\setlength{\parsep}{0pt}
\setlength{\parskip}{0pt}
\item{\textit{Task}}: A navigation instruction with a brief task description.
% \item{\textit{History}}: History representations at $t$-th step.
\item{\textit{Observation}}: Scene representations of all reachable viewpoints at the current location.
\item{\textit{Output Hint}}: \eg, {\it select a direction from the observation}.
\end{itemize}
The LLM takes the above schemas as input, to predict the ID of a viewpoint to move towards, where the ID is a number. As the agent moves, the history representations are updated with the scene representation corresponding to the agent's most recently selected viewpoint.

\para{Object Localization.} It requires identifying the correct object from a set of visible objects after the agent successfully reaches the destination. In addition to \textit{History} schema, it also contains the following schemas:
\begin{itemize}[leftmargin=*]
\setlength{\itemsep}{0pt}
\setlength{\parsep}{0pt}
\setlength{\parskip}{0pt}
\item{\textit{Task}}: An object localization command.
% \item{\textit{History}}: History representations until the last step.
\item{\textit{Observation}}: Object representations of all visible objects at the current position. The object representations are extracted from a pre-trained ViT and subsequently converted into the same dimension as word embeddings.
\item{\textit{Output Hint}}: \eg, {\it select an object from the observation}.
\end{itemize}
With these schemas, the agent is required to generate the ID of the selected object.

\para{Trajectory Summarization.} 
We follow \cite{fried2018speaker} to include the task of synthesizing instructions from given trajectories. For this task, it shares the \textit{History} and \textit{Observation} schemas as VLN, where the \textit{History} schema is optional depending on the dataset. Besides the two schemas, we also include:
\begin{itemize}[leftmargin=*]
\setlength{\itemsep}{0pt}
\setlength{\parsep}{0pt}
\setlength{\parskip}{0pt}
\item{\textit{Task}}: A concise description of the style for summarizing, \eg, fine-grained and coarse-grained.
\item{\textit{Output Hint}}: \eg, {\it Summarize the above trajectory}.
\end{itemize}

\para{3D Question Answering (3D-QA)} asks the agent to answer a question in a 3D scene.
Different from the previous tasks, in this task, the \textit{History} schema is not required. The following schemas are provided for 3D-QA.
% We represent the 3D environment as a series of images from different viewpoints, and encode them into the scene representations $\{s_1\}_{i=1}^n$.
\begin{itemize}[leftmargin=*]
\setlength{\itemsep}{0pt}
\setlength{\parsep}{0pt}
\setlength{\parskip}{0pt}
\item{\textit{Task}}: A question about an indoor scene.
% \item{\textit{History}}: History representations until the last step.
\item{\textit{Observation}}: Scene representations of images from different positions. We also utilize the previous scene encoding to process the scene representations.
\item{\textit{Output Hint}}: \eg, {\it Answer the question based on the scene}.
\end{itemize}
The model demands to generate a textual answer based on the above schemas.

\para{Embodied Question Answering (EQA).}
The agent is asked to first navigate to the location referred by a question, and then respond to the question accordingly.
We utilize the schemas of VLN and 3D-QA in two stages, respectively.

\section{Experimental Setup}
In this section, we first introduce the implementation details (\S \ref{sec:implementation}). Then we describe the evaluation datasets, metrics, and baseline methods for VLN (\S \ref{sec:exp_vln}), 3D-QA (\S \ref{sec:exp_3dqa}) and EQA (\S \ref{sec:exp_eqa}), respectively.

\subsection{Implementation Details}
\label{sec:implementation}

\para{Model Details.} We fine-tune the multi-view fusion module and LLM, where the former consists of a 2-layer transformer encoder with a hidden size of 1024 and the LLM is built upon Vicuna-7B-v0 \cite{peng2023instruction}. The ViT in the scene encoder is EVA-CLIP-02-Large (428M) \cite{EVA-CLIP} and is kept frozen during the training phase.
In addition, we leverage the object features extracted from ViT-B16 by Chen \textit{et al.} \cite{chen2022learning}.
% while the object features used for object grounding are extracted from ViT-B16 following \cite{chen2022learning}.

\para{Training Details.} We follow the previous works \cite{Chen_2022_CVPR, Gao_2023_CVPR, Chen_2022_HM3D_AutoVLN} to employ a two-stage training strategy.
Throughout both stages, we utilize the Adam optimizer with a learning rate of 3e-5. The model is trained for 10,000 steps in the pre-training stage and 5,000 steps in the multi-task fine-tuning stage with a batch size of 64.
It takes approximately 80 hours with 8 Nvidia A100 GPUs.
During the testing phase, we employ a sampling strategy with a temperature of 0.01 for action generation in the SOON and REVERIE tasks, to encourage more exploration. For other tasks, we opt for a greedy strategy in generating actions.

% We follow previous works [11, 31] to set up the hyperparameters, i.e., the view number $n$ is set to 36 for all tasks, and the history round $t$ is set to 15, 15, 30, 20 for R2R, REVERIE, CVDN and SOON, respectively.

\para{Training Data.} In the pre-training stage, we perform teacher-forcing training on the combined dataset from CVDN, SOON, R2R, REVERIE, ScanQA, and augmented data from R2R and REVERIE. 
In the multi-task fine-tuning stage, we alternate between teacher forcing and student forcing on the combined dataset from CVDN, SOON, R2R, REVERIE, ScanQA, and LLaVA-23k~\cite{liu2023visual}.

For object localization, we utilize the corresponding annotations from REVERIE and SOON.
For trajectory summarization, we convert the instruction-trajectory pairs of VLN datasets into trajectory-instruction pairs, where the trajectory serves as input and the instruction as output. 
As for 3D-QA, in addition to ScanQA, we also construct question-answer pairs from the fine-grained annotations on R2R~\cite{hong2020sub}.
In concrete, the constructed questions ask the model to predict the corresponding sub-instruction for a selected viewpoint.
Additionally, we held out the task of EQA to verify the generalization capability on out-of-domain tasks.

% \subsection{Datasets}
% \label{sec:datasets}
% We evaluate our model on four VLN datasets and a 3D question-answering dataset, ScanQA \cite{azuma2022scanqa}.
% The four VLN datasets, including CVDN \cite{thomason2020vision}, SOON \cite{zhu2021soon}, R2R \cite{anderson2018vision}, REVERIE \cite{qi2020reverie}, focus on navigation tasks based on instructions of different granularities and forms. 
% These datasets are split into train, val-seen, val-unseen, and test sets according to environments.
% ScanQA \cite{azuma2022scanqa} asks the agent to respond to a question about a 3D scene.
% The dataset is divided into train, val, and test sets.
% In addition, we also test our model on MP3D-EQA \cite{wijmans2019embodied} in a zero-shot setting.
% Since there are some data points with inaccurate endpoints, we perform calibration on the endpoint positions in the test set.
\subsection{Setup for VLN} 
\label{sec:exp_vln}

\para{Datasets.} We adopt four datasets, each addressing distinct challenges posed by VLN. These datasets are split into train, val-seen, val-unseen, and test sets according to environments.

\begin{itemize}[leftmargin=*]
\setlength{\itemsep}{0pt}
\setlength{\parsep}{0pt}
\setlength{\parskip}{0pt}
\item{\textbf{CVDN}~\cite{thomason2020vision}} requires the agent to navigate towards the target based on a dialog history, thereby requiring the ability to comprehend the dialog and interpret it as actions.

\item{\textbf{SOON}~\cite{zhu2021soon}} asks the agent to locate a thoroughly described object, which necessitates intricate alignment between rich semantic descriptions and the corresponding visual cues.

\item{\textbf{R2R}~\cite{anderson2018vision}} demands the agent to navigate following a step-by-step instruction.
To make effective decisions, it requires the agent to dynamically track the progress, demanding fine-grained alignment between history and instructions.

\item{\textbf{REVERIE}~\cite{qi2020reverie}} requires the agent to localize a distant target object according to a concise high-level instruction.

\end{itemize}

\para{Metrics.} 
We follow \cite{anderson2018vision} to evaluate our method on the following metrics: 
1) Sucess Rate (\textbf{SR}), whether the agent successfully reaches the target location within a predefined distance threshold. 
2) Success Rate Weighted by Path Length (\textbf{SPL}), calculated as the SR weighted by the ratio of the ground truth length and actual path length. 
3) Oracle Success Rate (\textbf{OSR}), SR given the oracle stop strategy.
4) Trajectory Length (\textbf{TL}), the overall distance covered by the agent during navigation. 
5) Goal Process (\textbf{GP}), the progress in meters towards the goal.
GP is adopted for the CVDN dataset, while SPL is employed as the primary evaluation metric for other datasets.

\para{Baseline Methods.} We compare our method with the latest SoTA methods on the CVDN, SOON, R2R, and REVERIE datasets. We do not consider methods with pre-exploration (e.g., AuxRN \cite{zhu2020vision}, RREx-BoT \cite{sigurdsson2023rrex}), and models augmented by new environments (e.g., HM3D-AutoVLN \cite{Chen_2022_HM3D_AutoVLN}).

\subsection{Setup for 3D-QA} 
\label{sec:exp_3dqa}
\para{Dataset.}
ScanQA dataset \cite{azuma2022scanqa} is a widely used dataset for 3D-QA, which is divided into train, val, and test sets. Here, we use val and `test w/ objects' sets for comparison. 
For the results on `test w/o objects' set, please refer to the appendix.

\para{Metrics.} We follow \cite{hong20233dllm} to evaluate our method with Exact Match (\textbf{EM}), METEOR, ROUGE-L, CIDER, and BLEU-4. 

\para{Baseline Methods.} We include some representative methods for comparison, including VoteNet+MCAN \cite{yu2019mcan}, ScanRefer+MCAN \cite{yu2019mcan}, and 3D-LLM \cite{hong20233dllm}. 3D-LLM is the current SoTA method on the ScanQA benchmark.

\subsection{Setup for EQA} 
\label{sec:exp_eqa}

\para{Dataset.} 
We test the zero-shot inference capability on the val split of the MP3D-EQA~\cite{wijmans2019embodied} dataset.
Since MP3D-EQA is generated from functional programs, there are some data with inaccurate endpoints. Therefore, we manually check the dataset and filter out those invalid data in our experiments.

% Since it is generated from functional programs, there are some data points with inaccurate endpoints, we perform calibration on the endpoint positions in the val set.

\para{Metrics.} We report SR and SPL for the navigation phase, and Accuracy (\textbf{ACC}) for the question-answering phase.

\para{Baseline Methods.} 
% We test the zero-shot inference performance on MP3D-EQA of DUET models, each separately trained on R2R, REVERIE, and SOON. In addition, we also compare our zero-shot model with the fully-supervised VQA model.
We compare our \modelname~with the fully-supervised VQA model~\cite{embodiedqa} and zero-shot DUET models~\cite{Chen_2022_CVPR} separately trained on R2R, REVERIE, and SOON.

% \subsection{Evaluation Metrics}
% \label{sec:metrics}
% \para{VLN} We follow \cite{anderson2018vision} to evaluate our method on the following metrics: 
% 1) Sucess Rate (\textbf{SR}), whether the agent successfully reaches the goal within a predefined distance threshold. 
% 2) Success Rate Weighted by Path Length (\textbf{SPL}), calculated as the SR weighted by the ratio of the ground truth length and actual path length. 
% % 3) Oracle Success Rate (\textbf{OSR}), SR given the oracle stop strategy.
% 3) Trajectory Length (\textbf{TL}), the overall distance covered by the agent during navigation. 
% 4) Goal Process (\textbf{GP}), the progress in meters towards the goal.
% SPL is employed as the primary evaluation metric.

% \para{3D-QA} We follow \cite{hong20233dllm} to evaluate our method with Exact Match (\textbf{EM}), METEOR, ROUGE-L, CIDER, and BLEU-4. 

% \para{EQA} We report TL and SPL for navigation, and Accuracy (\textbf{ACC}) for question-answering.

\begin{table*}[]
\centering
\renewcommand\arraystretch{1}
\setlength{\tabcolsep}{3.2mm}{
\begin{tabular}{l|cc|cc|cc|cc|cc} \toprule
\multicolumn{1}{c}{} & \multicolumn{2}{c}{\textbf{CVDN}} & \multicolumn{2}{c}{\textbf{SOON}} & \multicolumn{2}{c}{\textbf{R2R}} & \multicolumn{2}{c}{\textbf{REVERIE}} & \multicolumn{2}{c}{\textbf{ScanQA}} \\ 
\multicolumn{1}{c}{} & Val-U & \multicolumn{1}{c}{Test} & Val-U & \multicolumn{1}{c}{Test}  & Val-U & \multicolumn{1}{c}{Test} & Val-U & Test & \multicolumn{1}{c}{Val} & Test \\ \midrule
\multicolumn{11}{c}{\textit{\textbf{Separate Model For Each Task}}} \\ \hline
PREVALENT \cite{hao2020towards} & 3.15 & 2.44 & - & -  & 53 & 51 & - & -  & - & - \\
HOP \cite{Qiao2022HOP} & 4.41 & 3.24 & - & & 57 & 59 & 26.11 & 24.34 & - & - \\
HAMT \cite{NEURIPS2021_2e5c2cb8} & 5.13 & 5.58  & - & -  & 61 & \underline{60} & 30.20 & 26.67 & - & - \\
VLN-BERT \cite{Hong_2021_CVPR}& - & - & - & -  & 57 & 57 & 24.90 & 23.99 & - & -\\
GBE \cite{zhu2021soon} & - & - & 13.34 & 9.23 & - & - & - & - & - & - \\
DUET \cite{Chen_2022_CVPR}& - & - & 22.58 & \underline{21.42}  & 60 & 58 & 33.73 & \underline{36.06} & - & - \\
Meta-Explore \cite{Hwang_2023_CVPR} & - & - & - & \underline{25.80}  & 62 & \textbf{61} & 34.03 & - & - & - \\
AZHP \cite{Gao_2023_CVPR} & - & - & - & - & 61 & \underline{60} & \textbf{36.63} & 35.85 & - & -\\
VLN-SIG \cite{li2023vlnsig} & 5.52 & 5.83 &- & - & \underline{62} & \underline{60} & - & - & - & - \\
VLN-PETL \cite{qiao2023vln} & \underline{5.69} & \underline{6.13} & - & - & 60 & 58 & 27.67 & 26.73 & - & -\\
BEV-BERT \cite{an2023bevbert} & & & & & \textbf{64} & \underline{60} & \underline{36.37} & \textbf{36.41} & - & - \\
3D-LLM \cite{hong20233dllm} & - & - & - & - & - & - & - & - & \underline{20.5} & \underline{19.1} \\ \hline
\multicolumn{11}{c}{\textit{\textbf{Unified Model For All Tasks}}} \\ \hline
MT-RCM+Env \cite{wang2020environment} & 4.65 & \multicolumn{1}{c}{3.91} & - & \multicolumn{1}{c}{-}  & 49 & \multicolumn{1}{c}{40} & - & \multicolumn{1}{c}{-} & - & - \\
% \modelname\ v1 (Ours) & &  & \textbf{28.8} & \textbf{26.5} & \underline{61} & \underline{60} & \underline{35.61} & 32.33 & \textbf{21.5 } &\textbf{24.0} \\ 
% \modelname\ v2 (Ours) & 59 & \textbf{54÷} & \textbf{26.2} & 35.16 & \textbf{23.6} \\ 
% \modelname\ v3 & \textbf{6.16} & \textbf{7.51} & \textbf{28.22} & \textbf{25.95} & 60 & \textbf{61} & 33.39 & 31.05 & \textbf{22.1} & \textbf{24.5} \\
\modelname\ & \textbf{6.16} & \multicolumn{1}{c}{\textbf{7.90}} & \textbf{29.24} & \multicolumn{1}{c}{\textbf{26.26}} & 59 & \multicolumn{1}{c}{\underline{60}} & 35.68 & \multicolumn{1}{c}{32.33} & \textbf{23.0} & \textbf{26.3} \\
\bottomrule
\end{tabular}}
\vspace{-0.1cm}
\caption{Overall comparison with state-of-the-art methods on all tasks. `Val-U' denotes val-unseen split.
We report SPL for CVDN, SOON, R2R, and REVERIE, and report Accuracy for ScanQA.}
\vspace{-0.1cm}
\label{table:performance}
\end{table*}

\begin{table*}[t]
\centering
\renewcommand\arraystretch{1}
\resizebox{0.99\textwidth}{!}{
\setlength{\tabcolsep}{1.5mm}{
\begin{tabular}{l|ccccc|ccccc} \toprule
& \multicolumn{5}{c|}{ScanQA-Val} & \multicolumn{5}{c}{ScanQA-Test} \\
& EM & ROUGE-L & METEOR & CIDER & BLEU-4 &  EM & ROUGE-L & METEOR & CIDER & BLEU-4 \\ \midrule
VoteNet+MCAN \cite{yu2019mcan} & 17.3 & 29.8 & 11.4 & 54.7 & 6.2 & 19.7 & 30.9  & 12.0& 58.2 & 6.0 \\
ScanRefer+MCAN \cite{yu2019mcan} & 18.6 & 30  & 11.5& 55.4 & 7.9 & 20.6  & 30.7 & 11.9 & 57.4 & 7.5 \\
ScanQA \cite{azuma2022scanqa} & 21.0 & 33.3  & 13.1& 64.9 & 10.1 & 23.5  & 34.3 & 13.5& 67.3 & 12.0 \\ 
3D-LLM (flamingo) \cite{hong20233dllm} & 20.4 & 32.3  & 12.2& 59.2 & 7.2 & 23.2& 34.8  & 13.5 & 65.6 & 8.4 \\
3D-LLM (BLIP2-flant5) \cite{hong20233dllm} & 20.5 & 35.7 & 14.5& 69.4 & 12.0 & 19.1 & 35.3  & 14.9 & 69.6 & 11.6 \\

\textbf{\modelname} (Ours) & \textbf{23.0} & \textbf{38.4} & \textbf{15.4} & \textbf{75.9} & \textbf{12.5}  & \textbf{26.3} & \textbf{40.2} & \textbf{16.6} & \textbf{80.8} & \textbf{13.9} \\
\bottomrule
\end{tabular}}
}
\caption{Detail comparison with state-of-the-art methods on ScanQA.}
\vspace{-0.2cm}
\label{table:scanqa_sota}
\end{table*}

\section{Experimental Results}
\label{sec:experiments}
We conduct a series of experiments to answer three critical questions about \modelname: 
(1) Can \modelname, when trained with diverse tasks, demonstrate superior performance compared to existing SoTA methods (\S \ref{sec:performance})?
(2) How well does \modelname~generalize to unseen tasks, compared to previous task-specific models (\S \ref{sec:generalization})?
(3) What is the impact of each component in our method (\S \ref{sec:ablation})?
Lastly, we also provide visualization for \modelname~on unseen scenes and tasks (\S \ref{sec:visualization}).

\subsection{Comparision with SoTA Methods}
\label{sec:performance}

\para{Delivering SoTA Results with a Single Model.}
We present the comparison across all tasks in Table \ref{table:performance}.
Our single model achieves SoTA performance on the test sets of the CVDN, SOON, and ScanQA datasets, and demonstrates comparable results to the latest SoTA methods on R2R and REVERIE.

\para{Significant Improvement on CVDN Can Be Credited to Our Innovative Design.} 
Compared to the 6.13 GP of VLN-PETL \cite{qiao2023vln}, our method shows a significant increase in the GP at 7.90, winning the first place on the leaderboard of CVDN.
Compared to other datasets, the improvement on CVDN is the most pronounced. 
We attribute this significant improvement primarily to two factors: 1) The dialog structure in CVDN is relatively complex, and the knowledge inherited from LLM in our model can better help comprehend the dialog. 2) Given that the size of the CVDN dataset is smaller compared to other datasets, the unification of these datasets effectively mitigates the issue of data scarcity.

\para{Our Method Also Excels in 3D Tasks.} As shown in Table \ref{table:scanqa_sota}, our method achieves SoTA results on the val and test sets in all metrics.
It obtains a 26.3\% EM, with a 7.2\% improvement over 3D-LLM, which is specially designed for 3D tasks.

\para{Better Performance on Tasks with Complex Instructions.}
We count the average length of instructions across different datasets, with CVDN averaging 81.6 words per instruction, SOON averaging 38.6 words, R2R averaging 29 words, and REVERIE averaging 18 words. 
This reflects the complexity of the instructions to some extent.
We notice that our method exhibits superior performance on datasets with complex instructions, such as CVDN, SOON, and R2R, achieving performance better than or comparable to SoTA methods.
However, there is still a slight gap with DUET on datasets with relatively simple instructions, such as REVERIE.
This may indicate that our method possesses excellent instruction comprehension capabilities, which helps improve the performance on tasks with complex instructions.

\para{\textit{\modelname} Demonstrates An Excellent Object Localization Capability.} 
In the object localization task of REVERIE, our method achieves 19.83\% RGS and 16.04\% RGSPL, and outperforms the 14.88\% and 13.08\% achieved by HAMT \cite{NEURIPS2021_2e5c2cb8}, demonstrating an excellent object localization capability. 
Given that existing methods typically integrate object features with image features, we believe that our method could be further enhanced by combining these features.

\begin{table*}[]
\centering
\renewcommand\arraystretch{1}
\setlength{\tabcolsep}{3mm}{
\begin{tabular}{l|cc|cccc|cccc} \toprule
& \multicolumn{2}{c|}{CVDN}  & \multicolumn{4}{c|}{SOON}  & \multicolumn{4}{c}{REVERIE} \\ 
& TL & \textbf{GP$\uparrow$} & TL & \textbf{OSR$\uparrow$} & \textbf{SR$\uparrow$} & \textbf{SPL$\uparrow$} & TL & \textbf{OSR$\uparrow$} & \textbf{SR$\uparrow$} & \textbf{SPL$\uparrow$} \\ \midrule
DUET (R2R)  & 21.12 & 3.38 & 26.83 & 7.64 & 4.66 & 2.84 & 7.88 & 29.11 & 24.91 & 20.00  \\
DUET (REVERIE) & 76.13 & 3.30 & 33.72 & 20.86 & 10.24 & 6.06 & - & - & - & -  \\
DUET (SOON) & 48.61 & 2.40 & - & - & - & - & 38.10 & 43.45 & 10.91 & 3.64 \\
\textbf{\modelname} & 26.37  &\textbf{4.46} &28.66 & \textbf{33.11} & \textbf{19.81} & \textbf{14.29} & 18.96 & \textbf{51.47} & \textbf{28.10} & \textbf{21.04}  \\
\bottomrule
\end{tabular}}
\vspace{-0.1cm}
\caption{Held-out results on val-unseen splits of CVDN, SOON and REVERIE. We only perform the multi-task fine-tuning for held-out experiments.
Trajectory Length (TL) serves as a statistical indicator rather than an evaluation metric.
}
\vspace{-0.2cm}
\label{table:held_out}
\end{table*}

\begin{table}[]
\centering
\renewcommand\arraystretch{1.}
\setlength{\tabcolsep}{0.8mm}{
\resizebox{0.42\textwidth}{!}{
\begin{tabular}{l|lc|cccc} \toprule
\multirow{2}{*}{\#}& & \multirow{2}{*}{GT Path} & \multicolumn{3}{c}{Navigation} & \multicolumn{1}{c}{QA} \\
& & &TL & \textbf{SR$\uparrow$} & \textbf{SPL$\uparrow$} & \textbf{ACC$\uparrow$}  \\ \midrule
% \multicolumn{4}{c}{\textit{Navigation Trajectory}} \\ \hline
1 & DUET (R2R)~\cite{Chen_2022_CVPR} &  & 16.47 & 47.00 & 30.51 & - \\
2 & DUET (REVERIE)~\cite{Chen_2022_CVPR} & & 12.59 & 39.22 & 11.47 & -\\
3 & DUET (SOON)~\cite{Chen_2022_CVPR} & & 47.01 & 17.43 & 3.91 & -\\
% 4 & EQA (habitat-lab)$\dagger$~\cite{embodiedqa} & & & & & \\ 
4 & \textbf{\modelname} (Ours) & & 14.15 & 47.78 & 35.60 & 44.5 \\  \hline
5 & EQA (habitat-lab)$\dagger$~\cite{embodiedqa} & \checkmark & - & - & - & 46.0 \\ 
6 & \textbf{\modelname} (Ours) & \checkmark & - & - & - & 47.4 \\ 
% \multicolumn{4}{c}{\textit{Ground truth Trajectory}} \\ \hline
\bottomrule
\end{tabular}
}
}
\vspace{-0.1cm}
\caption{Zero-shot inference results on MP3D-EQA. `GT Path' means using the ground truth trajectory for question answering. $\dagger$ indicates the method is finetuned on the training set of MP3D-EQA. 
Trajectory Length (TL) serves as a statistical indicator rather than an evaluation metric.
}
\vspace{-0.3cm}
\label{table:eqa}
\end{table}

\begin{table*}[]
\centering
\renewcommand\arraystretch{1}
\setlength{\tabcolsep}{3.5mm}{
\resizebox{0.88\textwidth}{!}{
\begin{tabular}{l|ccc|c|cc|cc|cc|cc} \toprule
\multirow{2}{*}{\textbf{\#}} & \multirow{2}{*}{\textbf{LLM}} & \multirow{2}{*}{\textbf{Multi-Task}} & \multirow{2}{*}{\textbf{Pretrain}}  & \multicolumn{1}{c|}{CVDN} & \multicolumn{2}{c|}{SOON}& \multicolumn{2}{c|}{R2R}  & \multicolumn{2}{c|}{REVERIE} & \multicolumn{2}{c}{ScanQA} \\
& & &  & GP & SR & SPL & SR & SPL& SR & SPL & EM & ROUGE-L \\ \midrule
1& \checkmark & & & 5.54 & 28.37 & 21.37 & 64 & 57  & 30.95 & 24.10 & 21.8 & 37.0 \\
2& & \checkmark & & 3.64 & 20.73 & 17.39  & 49 & 40 & 31.49 & 26.87 & 12.4 & 22.8 \\
3& \checkmark & \checkmark &  & 5.91 & 35.44 & 28.09 & \textbf{67} & 58  & \textbf{44.56} & \textbf{36.63} & \textbf{23.3} & \textbf{38.2}  \\
4& \checkmark & \checkmark & \checkmark & \textbf{6.16} & \textbf{38.33} & \textbf{29.24}  & \textbf{67} & \textbf{59} & 42.15 & 35.68 & 22.1 & 37.6 \\
\bottomrule
\end{tabular}}
}
\vspace{-0.1cm}
\caption{Ablation study of \modelname\ across all tasks.
`LLM', `Multi-Task', and `Pretrain' denote the utilization of pretrained LLM weights for initialization, the execution of multi-task learning, and the performance of pre-training, respectively.
The results reported are from the val-unseen splits for VLN tasks and the val split for ScanQA.}
\vspace{-0.2cm}
\label{table:ablation}
\end{table*}

% Yiwu: suggestion for table

% \begin{table*}[]
% \centering
% \renewcommand\arraystretch{1}
% \setlength{\tabcolsep}{2mm}{
% \begin{tabular}{l|cc|c|cc|cc|cc} \hline
% & \multicolumn{2}{c|}{R2R} & \multicolumn{1}{c|}{CVDN} & \multicolumn{2}{c|}{SOON} & \multicolumn{2}{c|}{REVERIE} & \multicolumn{2}{c}{ScanQA} \\
% & SR & SPL & GP & SR & SPL & SR & SPL & EM & ROUGE-L \\ \hline
% single-task & 64 & 57 & 5.54 & 28.3 & 21.4 & 31.0 & 24.1 & 21.8 & 37.0 \\
% % multi-task & 68 & 61 & 6.16 & 41.1 & 35.6 & 41.1 & 35.6 & 21.5 & 36.1 \\
% multi-task + pretrain & 68 & 60 & 6.16 & 41.1 & 35.6 & 41.1 & 35.6 & 22.1 & 37.6 \\
% \hline
% \end{tabular}}
% \caption{Comparison of our method using single-task and multi-task learning.}
% \label{table:multi_task}
% \end{table*}

\subsection{Generalization Ability on Unseen Tasks}
\label{sec:generalization}
We evaluate the zero-shot inference performance of our method on unseen tasks, and compare it with zero-shot DUET models (DUET (R2R), DUET (REVERIE), and DUET (SOON)), with each model being separately trained on its corresponding dataset.

\para{Generalize to Out-of-Domain VLN Tasks.}
We conduct held-out experiments to verify the generalization ability to out-of-domain VLN tasks. 
Specifically, we individually exclude CVDN, SOON, and REVERIE from the training set, train three separate models, and then test their zero-shot performance on the respective excluded datasets.
% Respectively, we move the CVDN, SOON, and REVERIE tasks out of our training set and then test zero-shot navigation performance. 
% We compare our method to DUET \cite{Chen_2022_CVPR} models separately trained on R2R, REVERIE, and SOON. 
DUET specially designs different hyper-parameters and pre-training schemes for each VLN task, giving the model strong in-domain navigation capabilities. However, such learned task-specific agents lack out-of-domain generation abilities. As illustrated in Table \ref{table:held_out}, our method significantly outperforms DUET on CVDN and SOON, improving 32\% of GP on CVDN and 136\% SPL on SOON, respectively. Since instructions in REVERIE are relatively simpler and similar to R2R, DUET (R2R) delivers an SR of 24.91\% on REVERIE, but we still achieve a better SR of 28.10\%. 
This demonstrates that our schema-based instruction and multi-task learning empower the out-of-domain generation abilities.
% We conduct held-out experiments to verify the generalization ability of our method by removing the corresponding dataset from the training dataset. 

\para{Skill Combination for EQA.}
% Yiwu: suggestion for this zero-shot experiments
We perform a zero-shot evaluation on MP3D-EQA to show that \modelname~can combine the learned navigation and question-answering ability to solve more complex tasks. 
We ask our agent to first execute the navigation process and then answer the question after reaching the goal.
% Table \ref{table:eqa} demonstrates the zero-shot performance of our method on the MP3D-EQA dataset. 
% Our model is designed to execute navigation prior to answering questions upon reaching the goal, achieving 34.8\% SPL and 43.5\% Accuracy. 
% As illustrated in Table \ref{table:eqa}, our model achieves 34.8\% SPL and 43.5\% Accuracy.
% Notably, this performance exceeds that of DUET (R2R) by 4.3\% in SPL (rows 1 vs. 4).
As illustrated in Table \ref{table:eqa}, our model achieves 47.78\% SR and 35.60\% SPL, surpassing DUET (R2R) by 5.1\% in SPL (row 1 vs. 4).
At the same time, it can also answer questions at a decent accuracy of 44.5\%, while the DUET models are incapable of performing question answering.
Moreover, when the ground truth trajectories are provided, our zero-shot model presents superior performance over the fully-supervised EQA model (rows 5 vs. 6).

\begin{figure*}[t] %H为当前位置，!htb为忽略美学标准，htbp为浮动图形
\centering %图片居中
\scalebox{1.08}[1.0]{
\includegraphics[width=0.83\textwidth]{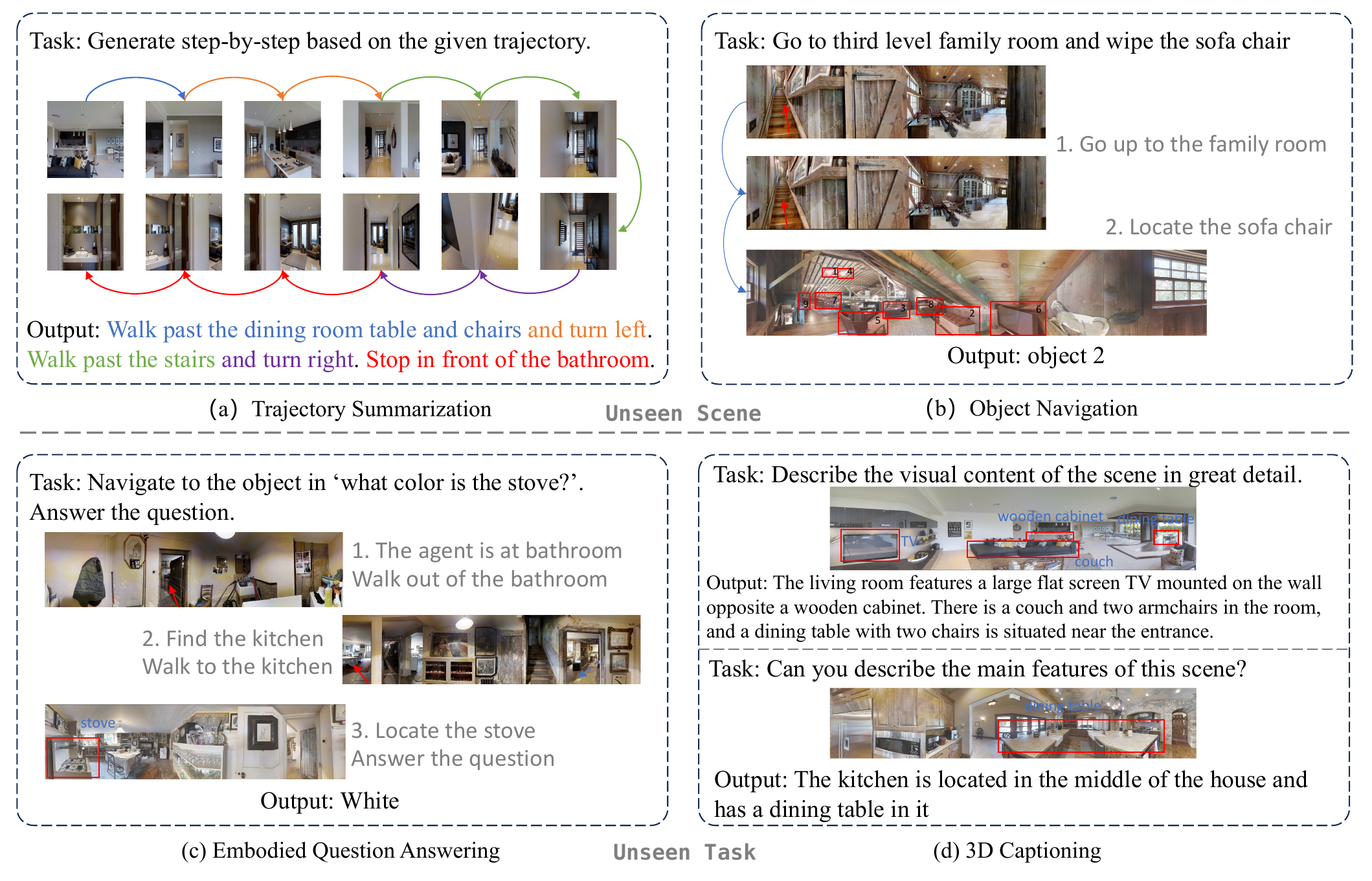} %插入图片，[]中设置图片大小，{}中是图片文件名
}
\vspace{-0.3cm}
\caption{The visualization for our method on unseen scenes and unseen tasks.
In Figure (a), lines and text of the same color represent sub-trajectories and their corresponding sub-instructions.
In Figures (b) and (c), the text in gray is the description of the actions of the agent during navigation, while the red arrow indicates the direction that the agent moves towards.}
\label{fig:visulaiaztion}
\vspace{-0.3cm}
\end{figure*}

\subsection{Ablation}
\label{sec:ablation}

\para{Multi-task Learning Enhances the Performance On All Tasks.}
Table \ref{table:ablation} illustrates that multi-task learning improves performance on all tasks (row 1 vs. 3).
This demonstrates that expanding the volume and diversity of training data is crucial for learning a generalist model for embodied navigation.

\para{LLM plays a Key Role in Our Method.}
Comparing rows 2 and 3, we can observe a significant performance drop when the LLM is randomly initialized, underscoring the substantial role that the LLM plays.

\para{Limited Benefits of Pre-Training on Augmented Data.}
Previous works \cite{Chen_2022_CVPR, Gao_2023_CVPR} have consistently shown notable improvements after pre-training on augmented data from R2R and REVERIE. 
However, comparing rows 3 and 4, we find only a slight enhancement on R2R, CVDN, and SOON after pre-training. 
We speculate that the quality of the data may play a more crucial role than its quantity for our method.

\subsection{Visualization}
\label{sec:visualization}
Figure \ref{fig:visulaiaztion} (a) and (b) are examples of trajectory summarization and object navigation on unseen scenes, respectively.
The first example illustrates that our model can generate accurate step-by-step instructions given trajectories, which could be further used for data augmentation.
Figure \ref{fig:visulaiaztion} (c) and (d) respectively present examples of EQA and 3D captioning, which are not encountered in training data, demonstrating the generalizability of \modelname.
Specifically, as shown in Figure \ref{fig:visulaiaztion} (d), our model is capable of producing captions of varying granularity according to the instructions.

\section{Conclusion}
\label{sec:conclusion}

In this paper, we present the first generalist model for embodied navigation, \modelname, which adapts LLMs to a variety of tasks by introducing schema-based instruction.
Benefiting from this design, we unify diverse tasks into a generation problem, allowing our model to utilize data sources from various datasets. Our experiments show that our single model can achieve SoTA results on CVDN, SOON, and ScanQA, and comparable performance to the latest models on R2R and REVERIE. 
Moreover, it also demonstrates strong generalizability and presents promising results on unseen tasks.

\section{Acknowledgements}
\label{sec:ack}
This work was supported by National Key R\&D Program of China (Project No. 2022ZD0161200, 2022ZD0161201). It was also partially funded by the Centre for Perceptual and Interactive Intelligence (CPIl) Ltd under the Innovation and Technology Commission (ITC)'s InnoHK. Liwei Wang is a Principal Investigator of CPII under the InnoHK. 
This work has also been supported by Hong Kong Research Grant Council - Early Career Scheme (Grant No. 24200223) as well as partially supported by Hong Kong Innovation and Technology Commission Project No. ITS/228/22FP.

{\small
\bibliographystyle{ieeenat_fullname}
\bibliography{11_references}
}

\ifarxiv \clearpage \appendix \maketitlesupplementary

\section{Dataset Statistics}
We provide training data statistics in Table \ref{table:statistics}. 

\para{CVDN.} In the CVDN dataset, the training section includes 4,742 questions spanned across 57 different environments. The unseen validation subset comprises 907 questions located in 10 unique environments. Meanwhile, the test subset consists of 1,384 questions over 16 environments.

\para{SOON} There are 2,780 instructions over 34 houses in the training set, while the unseen validation set hosts 339 instructions from 5 houses. The test set, on the other hand, contains 1,411 instructions derived from 14 houses.

\para{R2R} is comprised of a training set with 14,039 instructions drawn from 61 unique houses, a validation unseen set (val-unseen) containing 2,349 instructions from 11 houses, and a test set which includes 4,173 instructions from 18 houses.

\para{REVERIE.} For the REVERIE dataset, the training set encompasses 10,466 instructions across 60 houses, the unseen validation set (val-unseen) consists of 3,521 instructions from 10 houses, and the test set includes instructions from 16 houses, totaling 6,292.

\para{ScanQA.} The training set holds 25,563 questions spread across 562 scenes, the validation set consists of 4,675 questions within 71 scenes, and the test set without objects (test w/o objects) includes 6,179 questions from 97 scenes.

\para{EQA.} For EQA, we filter out the data in the validation set with inaccurate endpoints, retaining 849 entries for testing.

\section{Hyperparameter Details}
We follow previous works [11, 31] to set up the hyperparameters, i.e., the view number $n$ is set to 36 for all tasks, and the history round $t$ is set to 15, 15, 30, 20 for R2R, REVERIE, CVDN and SOON, respectively.

\section{Ablation for Schema Elements}
As shown in Table \ref{table:ablation_schema}, we observe a significant drop when removing these elements. This reveals the effectiveness of the proposed schemas in distinguishing different tasks.

\begin{table}[]
\centering
\renewcommand\arraystretch{1.}
\setlength{\tabcolsep}{0.8mm}{
\resizebox{0.5\textwidth}{!}{
\begin{tabular}{cc|ccccc|c} \toprule
O & T & CVDN  & SOON & R2R & REVERIE & ScanQA & Sum$\uparrow$ \\ \toprule
- & - & 3.51 & 20.68 & 37.51 & 23.93 & 21.86 & 107.49 \\
\checkmark & - & 5.07 & 24.52 & 50.29 & 28.27 & 21.50 & 129.65  \\
\checkmark & \checkmark & 5.40 & 26.32 & 55.07 & 30.56 & 23.10 & 140.45 \\
\bottomrule
\end{tabular}
}
}
\caption{The ablation of schemas. 
We train all models for 2,500 steps and report the results on validation unseen sets. $O$: output hint. $T$: task hint. 
}
\label{table:ablation_schema}
\end{table}

\section{Detailed Comparison on Benchmarks}
We provide the detailed comparison of CVDN, SOON, R2R, and REVERIE in Table \ref{table:cvdn_sota}, \ref{table:soon_sota}, \ref{table:r2r_sota}, and \ref{table:reverie_sota}, respectively. 
We present the results of the top 5 teams on the ScanQA leaderboard up to Nov. 2023 in Table \ref{table:scanqa_board}, and our method wins second place on the leaderboard.

\section{Instruction Templates}
Based on the proposed schema-based instruction, we present the instruction templates as shown in Table \ref{table:instruction}.

\begin{table}[]
\centering
\renewcommand\arraystretch{1}
\setlength{\tabcolsep}{5mm}{
\begin{tabular}{l|cc} \hline
& Val-Unseen & Test \\ \hline
Seq2Seq \cite{thomason2020vision} & 2.10 & 2.35 \\
PREVALENT \cite{hao2020towards} & 3.15 & 2.44 \\
HOP \cite{Qiao2022HOP} & 4.41 & 3.31 \\
MT-RCM \cite{wang2020environment} & 4.36 & - \\
MT-RCM+Env \cite{wang2020environment} & 4.65 & 3.91 \\
HAMT \cite{NEURIPS2021_2e5c2cb8} & 5.13 & 5.58 \\
VLN-SIG \cite{li2023vlnsig} & 5.52 & 5.83 \\
VLN-PETL \cite{qiao2023vln} & 5.69 & 6.13 \\
\hline
\modelname~(Ours) & 6.16 & 7.90 \\
\bottomrule
\end{tabular}}
\caption{Detailed comparison with state-of-the-art methods on CVDN.}
\label{table:cvdn_sota}
\end{table}

\begin{table*}[]
\centering
\renewcommand\arraystretch{1.}
\setlength{\tabcolsep}{3.5mm}{
\resizebox{0.8\textwidth}{!}{
\begin{tabular}{l|cccccc|cc} \toprule
& CVDN  & SOON & R2R & REVERIE & ScanQA & LLaVA & R2R$\dagger$ & REVERIE$\dagger$ \\ \toprule
scene & 57 & 34 & 61 & 60 & 562 & - & 60 & 60 \\
inst. & 4.7k & 27.8k & 14k & 10.5k & 25.6k & 23k & 1070k & 20k \\
\bottomrule
\end{tabular}
}
}
\caption{Training data statistics. \textbf{scene} and \textbf{inst.} indicate the number of 3D scenes and instances, respectively.
$\dagger$ means augmentation datasets provided by [11].
}
\label{table:statistics}
\end{table*}

\begin{table*}[]
\centering
\renewcommand\arraystretch{1}
\resizebox{0.8\textwidth}{!}{
\setlength{\tabcolsep}{3mm}{
\begin{tabular}{l|cccc|cccc} \toprule
& \multicolumn{4}{c|}{Val-Unseen} & \multicolumn{4}{c}{Test Unseen}\\ 
& TL & \textbf{OSR}$\uparrow$ & \textbf{SR}$\uparrow$ & \textbf{SPL}$\uparrow$ & TL & OSR$\uparrow$ & SR$\uparrow$ & SPL$\uparrow$\\ \midrule
GBE \cite{zhu2021soon} & 28.96 & 28.54 & 19.52 & 13.34  & 21.45 & 28.96 & 19.52 & 13.34 \\ 
DUET \cite{Chen_2022_CVPR} & 36.20 & 50.91 &  36.28 & 22.58 & 41.83 &   43.00 & 33.44 & 21.42 \\
AZHP \cite{Gao_2023_CVPR} & 39.33 &56.19 & 40.71 & 26.58 & - & - & - \\ 
Meta-Explore \cite{Hwang_2023_CVPR} & - & - & - & - & - & 48.7 & 39.1 & 25.8 \\ \hline
\modelname~(Ours) & 27.72 & 53.24 & 38.33 & 29.24 & 28.80 & 45.78 & 35.04 & 26.26 \\
\bottomrule
\end{tabular}}
}
\caption{Detailed comparison with state-of-the-art methods on the val-unseen split of the SOON dataset.}
\label{table:soon_sota}
\end{table*}

\begin{table*}[]
\centering
\renewcommand\arraystretch{1}
\resizebox{0.8\textwidth}{!}{
\setlength{\tabcolsep}{3.3mm}{
\begin{tabular}{l|cccc|cccc} \toprule
& \multicolumn{4}{c|}{Val Unseen} & \multicolumn{4}{c}{Test Unseen} \\
& TL & \textbf{NE}$\downarrow$ & \textbf{SR}$\uparrow$ & \textbf{SPL}$\uparrow$ & TL & \textbf{NE}$\downarrow$ & \textbf{SR}$\uparrow$ & \textbf{SPL}$\uparrow$  \\ \midrule
PREVALENT \cite{hao2020towards} & 10.19 & 4.71 & 58 & 53 & 10.51 & 5.30 & 54 & 51 \\
HOP \cite{Qiao2022HOP} &12.27 & 3.80 & 64 & 57 & 12.68 & 3.83 & 64 & 59 \\
HAMT \cite{NEURIPS2021_2e5c2cb8} & 11.46 & 2.29 & 66 & 61 & 12.27 & 3.93 & 65 & 60 \\
VLN-BERT \cite{Hong_2021_CVPR} & 12.01 & 3.93 & 63 & 57  & 12.35 & 4.09 & 63 & 57 \\
DUET \cite{Chen_2022_CVPR} & 13.94 & 3.31 & 72 & 60 & 14.73 & 3.65 & 69 & 59 \\
Meta-Explore \cite{Hwang_2023_CVPR}& 13.09 & 3.22 & 72 & 62 & 14.25 & 3.57 & 71 & 61 \\
AZHP \cite{Gao_2023_CVPR} & 14.05 & 3.15 & 72 & 61 & 14.95 & 3.52& 71 & 60 \\
VLN-SIG \cite{li2023vlnsig} & -& - & 72 & 62 & - & - & 72 & 60 \\ 
VLN-PETL \cite{qiao2023vln} & 11.52 & 3.53 & 65 & 60 & 12.30 & 4.10 & 63 & 58 \\  \hline
\modelname\ (Ours) & 12.81 & 3.51 & 67 & 59 & 13.21 & 3.71 & 68 & 60 \\
\bottomrule
\end{tabular}}
}
\caption{Detailed comparison with state-of-the-art methods on the R2R dataset. \textbf{NE} means navigation error.}
\label{table:r2r_sota}
\end{table*}

\begin{table*}[]
\centering
\renewcommand\arraystretch{1}
\resizebox{0.8\textwidth}{!}{
\setlength{\tabcolsep}{3.mm}{
\begin{tabular}{l|cccc|cccc} \hline
& \multicolumn{4}{|c|}{Val Unseen} & \multicolumn{4}{c}{Test} \\
& TL & \textbf{OSR}$\uparrow$ & \textbf{SR}$\uparrow$ & \textbf{SPL}$\uparrow$ & TL & \textbf{OSR}$\uparrow$ & \textbf{SR}$\uparrow$ & SPL$\uparrow$ \\ \hline
Seq2Seq \cite{anderson2018vision} & 11.07 & 8.07 & 4.20 & 2.84 & 10.89 & 6.88 & 3.99 & 3.09 \\
HOP \cite{Qiao2022HOP} & 18.85 & 36.24 & 31.78 & 26.11 & 16.38 & 33.06 & 30.17 & 24.34 \\
HAMT \cite{NEURIPS2021_2e5c2cb8} & 14.08 & 36.84 & 32.95 & 30.20 & 13.62 & 33.41 & 30.40 & 26.67 \\
VLN-BERT \cite{Hong_2021_CVPR} & 16.78 & 35.02 & 30.67 & 24.90 & 15.68 & 32.91 & 29.61 & 23.99 \\
DUET \cite{Chen_2022_CVPR} & 22.11 & 51.07 & 46.98 & 33.73 & 21.30 & 56.91 & 52.51 & 36.06 \\
AZHP \cite{Gao_2023_CVPR} & 22.32 & 53.65 & 48.31 & 36.63 & 21.84 & 55.31 & 51.57 & 35.85 \\
VLN-PETL \cite{qiao2023vln} & 14.47 & 37.03 & 31.81 & 27.67 & 14.00 & 36.06 & 30.83 & 26.73 \\ \hline
\modelname & 15.34 & 52.27  & 42.15 & 35.68 & 15.16 & 51.75 & 39.80 & 32.33 \\ 
\modelname w/p pretrain & 16.04 & 53.74 & 44.56 & 36.63 & 16.39 & 56.21 & 43.49 & 34.45\\ \hline
\end{tabular}}
}

\caption{Experimental results on REVERIE val-unseen and test sets.}
\label{table:reverie_sota}
\end{table*}

\begin{table*}[]
\centering
\renewcommand\arraystretch{1}
\resizebox{0.8\textwidth}{!}{
\setlength{\tabcolsep}{2.5mm}{
\begin{tabular}{l|l|cccccc} \hline
Rank&Team&\textbf{EM}$\uparrow$&\textbf{BLEU-1}$\uparrow$ & \textbf{BLEU-4}$\uparrow$ & \textbf{ROUGE}$\uparrow$ & \textbf{METEOR}$\uparrow$ & \textbf{CIDEr}$\uparrow$ \\ \toprule
1&mare & 30.82&34.41&17.75&41.18&15.60&79.35 \\
2&\modelname~(Ours) & 24.77 &38.72&12.49&37.95&15.55&74.90 \\
3&bubble-bee&23.78&33.26&14.71&34.37&13.86&67.62\\ 
4&Optimus-prime&23.01&30.15&11.85&32.76&12.92&62.62 \\
5&ML&21.37&32.69&11.73&32.41&13.28&62.8 \\
\bottomrule
\end{tabular}}
}

\caption{Top 5 Teams on the ScanQA Leaderboard as of Nov 2023}
\label{table:scanqa_board}
\end{table*}

\begin{table*}
\small
\begin{center}
\rowcolors{2}{white}{lightgray}
\begin{subfigure}[b]{1.0\textwidth}
         \centering
\begin{tabular}{|p{0.2\textwidth}| p{0.8\textwidth}}
\hline
\textbf{Dataset}  & \textbf{Example} \\ \hline\hline
CVDN & Find the described room according the given dialog. Target: The goal room contains a fireplace. Question: where should I go? Answer: If you turn around and keep heading then there is an open door on your right. Question: okay where to? Answer: Go through the open doors on the right. \\
SOON & Find the described target. Target: I want to find a ceramic, rectangular and white sink, which is set in the washroom. The sink is under the mirror and next to the toilet. The washroom is inside the office, which is on the first floor and next to the living room. \\
R2R & Navigate following the instruction. walk towards the faucet sculpture, continue past it on left side and walk to the right of the marble cylinder to the doorway. Walk through the doorway to the right of the long rug, around the foot of the bed, and through the doorway on the left. Stop in front of the towel rack.\\
REVERIE & Go to the location to complete the given task. Task: Proceed to the office and turn on the ceiling fan. \\
R2R-Summ & Predict the fine-grained instruction based on your previous history and current location. Fine-grained instructions contain commands for each individual step.\\
SOON-Summ & Generate the target you want to find based on your previous history and current location. Describe both the target and its surroundings. \\
REVERIE-Summ & Generate the task you need to complete based on your previous history and current location. \\
ScanQA & Please answer questions based on the observation. \\
REVERIE-OG, SOON-OG & Select the target object from the candidate objects based on the instruction and history. \\
\hline
\end{tabular}
\caption{Task}
\end{subfigure}

\begin{subfigure}[b]{1.0\textwidth}
         \centering
\begin{tabular}{|p{0.2\textwidth}| p{0.8\textwidth}}
\hline
\textbf{Dataset}  & \textbf{Example} \\ \hline\hline
Any Datasets & Following is the History, which contains the visual information of your previous decisions. \\ \hline
\end{tabular}
\caption{History}
\end{subfigure}

\begin{subfigure}[b]{1.0\textwidth}
         \centering
\begin{tabular}{|p{0.2\textwidth}| p{0.8\textwidth}}
\hline
\textbf{Dataset}  & \textbf{Example} \\ \hline\hline
\makecell[l]{ CVDN, SOON, \\R2R, REVERIE} & Following is the Candidate, which contains several directions you can go to at the current position, candidate (0) is stop. \\
\makecell[l]{R2R-Summ, SOON-Summ\\ REVERIE-Summ}
& Following is the Observation, which contains panoramic views at your current location. \\
REVERIE-OG, SOON-OG & Following is the Object, which contains several objects that you could see at the current viewpoint, option (0) indicates not exist.\\
ScanQA & The following is the Observation, which includes multiple images from different locations. \\
\hline
\hline
\end{tabular}
\caption{Observation}
\end{subfigure}

\begin{subfigure}[b]{1.0\textwidth}
         \centering
\begin{tabular}{|p{0.2\textwidth}| p{0.8\textwidth}}
\hline
\textbf{Dataset}  & \textbf{Example} \\ \hline\hline
CVDN & Understand the dialog in the Instruction and infer the current progress based on the History and dialog. Then select the correct direction from the candidates to go to the target location. \\
SOON & Nearby areas and objects can assist you in locating the desired room and object. Select the correct direction from the candidates to go to the target location. \\
R2R & Compare the History and Instruction to infer your current progress, and then select the correct direction from the candidates to go to the target location.\\
REVERIE & Explore the scene to find out the targeted room and object. Then select the correct direction from the candidates to go to the target location. \\
R2R-Summ & Please generate the step-by-step instruction. \\
SOON-Summ & Please predict both the target you want to find and its surroundings. \\
REVERIE-Summ & Please predict the task you need to complete. \\
REVERIE-OG, SOON-OG & Following is the Object, which contains several objects that you could see at the current viewpoint, option (0) indicates not exist.\\
\hline
\end{tabular}
\caption{Output Hint}
\end{subfigure}

\end{center}
\caption{All instruction templates used in our method.
R2R-Summ, SOON-Summ, and REVERIE-Summ are the trajectory summarization datasets created from R2R, SOON, and REVERIE, respectively. REVERIE-OG and SOON-OD are employed for object localization.}
\label{table:instruction}
\end{table*} \fi

\end{document}